\documentclass[10pt,twocolumn,letterpaper]{article}

\usepackage{iccv}
\usepackage{times}
\usepackage{epsfig}
\usepackage{graphicx}
\usepackage{amsmath}
\usepackage{amssymb}

\usepackage{subfig}
\usepackage{subfloat}
\usepackage{dblfloatfix}

\newcommand{\figref}[1]{\figurename~\ref{#1}}

\usepackage[breaklinks=true,bookmarks=false]{hyperref}

\iccvfinalcopy 

\def\iccvPaperID{****} 

\ificcvfinal\fi
\begin{document}

\title{Picking groups instead of samples: A close look at Static Pool-based Meta-Active Learning}

\author{Ignasi Mas\\
\small{Universitat Polit\`{e}cnica de Catalunya}\\
\small{Barcelona}\\
{\tt\small i.masmend@gmail.com}
\and
Josep Ramon Morros\\
\small{Universitat Polit\`{e}cnica de Catalunya}\\
\small{Barcelona}\\
{\tt\small ramon.morros@upc.edu}
\and
Ver\'{o}nica Vilaplana\\
\small{Universitat Polit\`{e}cnica de Catalunya}\\
\small{Barcelona}\\
{\tt\small veronica.vilaplana@upc.edu}
}

\maketitle
\ificcvfinal\thispagestyle{empty}\fi

\begin{abstract}
   \textit{Active Learning techniques are used to tackle learning problems where obtaining training labels is costly. In this work we use Meta-Active Learning to learn to select a subset of samples from a pool of unsupervised input for further annotation. This scenario is called Static Pool-based Meta-Active Learning. We propose to extend existing approaches by performing the selection in a manner that, unlike previous works, can handle the selection of each sample based on the whole selected subset.}
\end{abstract}

\section{Introduction}
\label{sec:intro}

In a standard supervised classification problem, the system learns from several labeled samples from each class during the training stage, in order to classify new samples at test time. However, annotated data is not always available. Active Learning (AL) suggests a scenario where a specific cost is associated to labeling. This is a common setting in cases such as a company that has to pay some employees for labeling data.

On the other hand, Few-Shot Learning \cite{fewshot:survey} suggests another scenario where the system is limited to train over a few samples per class. The extreme case is One-Shot Learning, where only one sample per class is available for training. There are several ways to approach the Few-Shot Learning task. Methods include getting more data (\eg multimodal learning~\cite{multimodal:survey}, domain adaptation~\cite{domainadaptation:survey} or data augmentation, depending on the available resources) or learning from data on similar problems.

Each Few-Shot problem can be solved using classical supervised learning: given a set of labeled data (the problem training set), the model is trained and it is expected to generalize on new samples. Later, the performance is evaluated with different samples of the same problem (its test set). We refer to this process as the Learning level.

We may want a system capable to adapt to a variety of problems for which it has not been specifically trained. One solution is to present the system several Few-Shot problems, different from the ones we will solve at test time but of similar nature. For each of these problems, annotated data should be available. We can train such a system by asking it to solve each one of these problems, analyze the results and update the model so that the error is minimized. This way, the system can generalize to new problems as it learns how to learn each problem. This process is named Meta-Learning because it is performed above the scope of a particular problem. At this level, there is a meta-training step (the process described before to adjust the model) and a meta-test step, to validate the complete system. The meta-test step consists of applying the system obtained in the meta-training step to a different set of problems (the meta-test set) and evaluating the results. The meta-test evaluation is the aggregation of all the single problem evaluations. We call the process where this pipeline works the Meta-Learning level.

The cross-over of Meta-Learning and Active Learning leads to Meta-Active Learning, which suggests the use of Meta-Learning to solve Active Learning problems at the Learning level. The AL component that performs the selection has to be trained as well. Clearly, the learning process involved here can not be performed at the Learning level (within the scope of a single problem) because the goal is to apply the automatic selection to new problems where no labeled data is available. Instead, a Meta-Leaning approach described in the above paragraph can be used to teach the AL component how to select the optimal samples to label.

The scenario where the system selects the samples from a pool of unlabelled data is called Pool-based Active Learning, in contrast with the case where samples arrive sequentially without any clue about future samples, that is named Stream-based Active Learning. One common problem with Pool-based Active Learning is that in the real world the system cannot always ask feedback without restrictions from the human annotation agent (called Oracle). This problem can be alleviated by using the so called Batch mode, where a batch of samples is selected before asking for labels, so that the selection does not require feedback from the Oracle for each sample. However, it still relies on a process where the Oracle is asked several times sequentially. Instead, we will focus on Static Pool-based Meta-Active Learning, where the system selects the whole subset of samples before asking for their labels. Thus, for each problem the Oracle is asked just once. More insight into the different scenarios is given in \ref{sbsec:sota_activelearning}.

The task of Static Pool-based Meta-Active Learning has already been studied by Contardo \etal in~\cite{DBLP:journals/corr/ContardoDA17}, where each sample is scored with a probability of being selected. Their approach allows learning a selection strategy that favors useful and representative data over deviant examples in a single step. However, it has some limitations which, depending on the scenario conditions, can be very harmful. We focus on the issue that each sample is selected independently (sample-focused), and in further sections we explain why it is an issue.

In our work, we propose a redefinition of the selection strategy to make it group-focused as opposed to sample-focused. This means that it should select each sample depending on the rest of the selected samples. This consideration should give the capacity to handle a final subsampling of the training data in a way that considers all possible groups of selection, so the probability is estimated per group instead of per sample.

We study the improvement of this method over the single step estimation. Furthermore, we provide results on the Omniglot dataset~\cite{dataset:omniglot} and reason about the performance of our approach on a computer vision challenge.

\section{Related work}
\label{sec:sota}

\subsection{Meta-Learning}
\label{sbsec:sota_metalearning}

Meta-Learning has been studied from many different approaches. Santoro \etal~\cite{DBLP:journals/corr/SantoroBBWL16} proposed to solve the One-Shot Learning problem from a Memory perspective, following the idea presented by Graves \etal~\cite{GravesWD14}. Later, the idea of metric learning introduced a new family of algorithms for Meta-Learning \cite{DBLP:journals/corr/VinyalsBLKW16,DBLP:journals/corr/SnellSZ17,Koch2015SiameseNN,10.1109/CVPR.2018.00131}. Another approach suggested to find a proper parameter initialization to generalize enough for all the problems, proposed by Finn \etal~\cite{DBLP:journals/corr/FinnAL17} and extended by several works \cite{DBLP:journals/corr/abs-1803-02999,Finn2018ProbabilisticMM,NIPS2018_7963}.  The last family of methods follows the idea of learning a proper optimizer \cite{DBLP:journals/corr/AndrychowiczDGH16,  DBLP:conf/iclr/RaviL17,DBLP:journals/corr/MishraRCA17}.

\subsection{Active Learning}
\label{sbsec:sota_activelearning}

All Active Learning problems share the property of having a cost assigned to the labels. With that in mind, we can define two scenarios.

The first one (Stream-based Active Learning) corresponds to the case where an agent receives data in an online manner (i.e. samples arrive consecutively and there is no clue about future samples) and has to decide either to label it or not. Many approaches have been proposed for solving this case \cite{querybycommittee, Roy2001TowardOA}.

The second scenario (Pool-based Active Learning) gives the agent access to all the (unsupervised) data at once. Depending on the specific conditions of the problem different subtypes of Pool-based Active Learning can be defined. On a Static scenario, the selection is made at once before asking the labels to the Oracle. On a Sequential scenario, the system gets feedback from the Oracle for part of the selected subset before continuing the selection. This difference is crucial since the second scenario is not always feasible, as stated in ~\cite{NIPS2012_4575}. Few Static Pool-based solutions have been proposed yet \cite{Yu:2006:ALV:1143844.1143980}. 

In Sequential Pool-based Active Learning, two modes can be defined: Single-instance mode \cite{Collet2014ActiveLF} or Batch mode \cite{Hoi2008,DBLP:journals/corr/abs-1902-06494}. In the first one each step is performed over a single sample while in the second one it is applied over a batch.

\subsection{Meta-Active Learning}
\label{sbsec:sota_metaactivelearning}

Meta-Active Learning still has few proposals. Most of them make use of Reinforcement Learning to guide the learning across Learning problems by exploring the performance when each given subset is selected. There are works for both the Stream-based scenario \cite{DBLP:journals/corr/WoodwardF17} and the Pool-based Meta-Active one in all its settings (specified in \ref{sec:intro}): Sequential Single-instance mode \cite{DBLP:journals/corr/abs-1708-00088}, Sequential Batch mode \cite{Ravi2018MetaLearningFB} and finally the Static scenario (where the subset is selected before labeling it). This last one was studied by Contardo \etal~\cite{DBLP:journals/corr/ContardoDA17}, and it is the focus of our work.

\section{Method}
\label{sec:method}

\subsection{Scenario}
\label{sbsec:scenario}

From now on we will differentiate between two levels, named Meta-Learning and Learning. Each Active Learning problem is solved at the Learning level (it learns from its available data), while the pipeline which updates the model is driven in the Meta-Learning level (through different Active Learning problems).

\begin{figure*}[ht!]
  \centering
  \includegraphics[width=16cm]{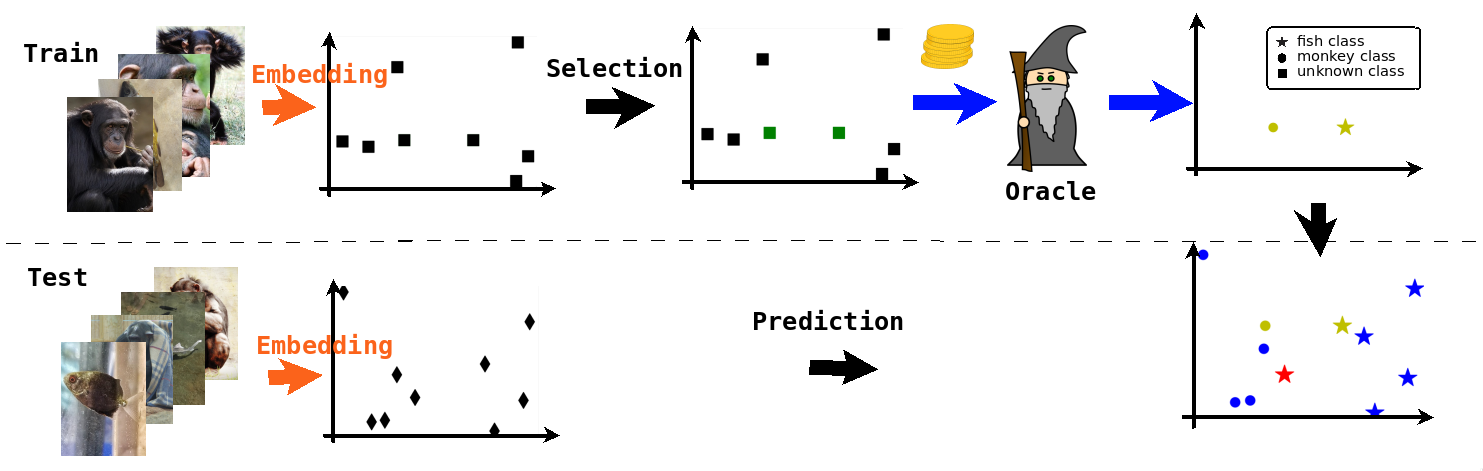}
  \caption {Example of a problem at the Learning level (binary classification between monkeys and fishes). Data samples are split into two sets, training and test, and embedded into an F-feature space ($F=2$). The budget allows picking two samples ($B = 2$) from the training set, which are sent to an Oracle that returns their true labels, creating the supervised training set. A prediction algorithm (1-NN classifier) is applied to the test set based on the supervised training set. Most of the predictions result in hits, in blue, and a single fail, in red.}
  \label{fig:learning_problem}
\end{figure*}

The detailed process of a single problem at the Learning level is illustrated in \figref{fig:learning_problem}, which uses as an example a binary classification problem between monkeys and fishes. 

A generic problem $p_{i}$ is specified as follows:

\begin{itemize}
  \item $K$ classes to classify.
  \item $N$ samples belonging to any of the $K$ classes ($N$ samples in total, without restrictions on how they are distributed along classes), which build the labeled training set $S_{\text{train}}$ and its unlabeled version $U_{\text{train}}$.
  \item $M$ samples belonging to any of the $K$ classes, that build the labeled set $S_{\text{test}}$, which we split between its unlabeled set $U_{\text{test}}$ and its labels $Y_{\text{test}}$ (for evaluation).
  \item A given budget $B$ that determines the number of samples that we will be able to label from the unlabeled training set $U_{\text{train}}$, to get a labeled subset $D$.
\end{itemize}

As \figref{fig:learning_problem} shows, to solve an Active Learning problem the process below is followed:

\begin{enumerate}
    \item Select $B$ samples from the represented $U_{\text{train}}$ set and send them to the Oracle who should return the labeled subset $D$.
    \item Predict classes for all samples in $U_{\text{test}}$ based on $D$, giving $\hat Y_{\text{test}}$.
    \item Evaluate $\hat Y_{\text{test}}$ against $Y_{\text{test}}$ (true labels).
\end{enumerate}

\begin{figure*}[ht]
  \centering
  \medskip
  \subfloat[Iterative pipeline]{\includegraphics[width=0.23\textwidth]{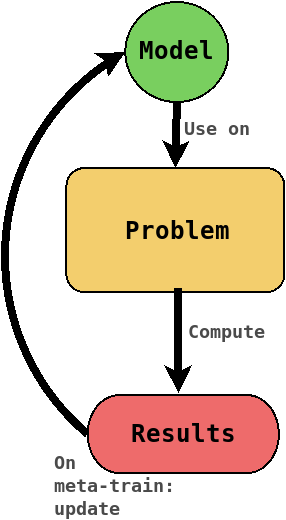}}
  \hfill
  \subfloat[Unrolled version of the pipeline]{\includegraphics[width=0.7\textwidth]{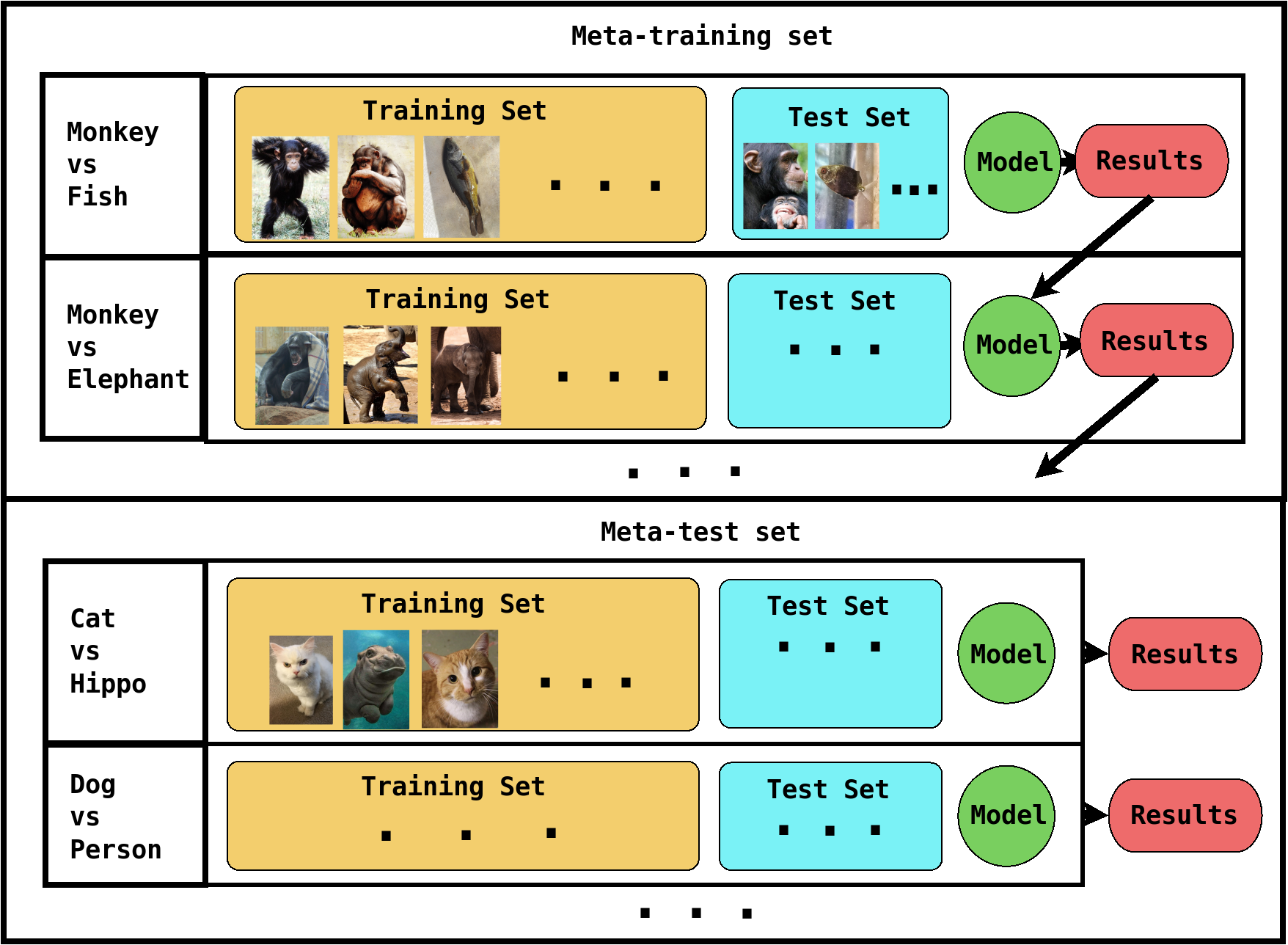}}
  \caption{Pipeline followed to guide the Meta-Learning process. At each epoch, a problem like the one represented in \figref{fig:learning_problem} is presented and the model updates based on the results it gets. Note that at each epoch the presented problem has, for both the training and test sets, samples from a different combination of classes, while within the problem both sets have different samples from the same combination of classes. Furthermore, there is no class overlapping between the classes in the Meta-training and Meta-test (meta-)sets.}
  \label{fig:metalearning_pipeline}
\end{figure*}

The pipeline at the Meta-Learning level is depicted in \figref{fig:metalearning_pipeline}. In order to build this pipeline, we need a  set of classes $C$, where for each class a pool of samples and its associated labels are available. Ideally, at a high level, the classes should be uniformly distributed according to what is expected at production (\eg if we expect to work on problems of classification of animals in general, classes for different animals should be used, trying to cover all the diversity of animals, instead of classes of animals and classes of cars).

The idea is to build a pipeline where several Learning problems guide the (meta-)training of the algorithm and allow a (meta-)evaluation of it. The process is driven through the following steps:

\begin{enumerate}
    \item Split classes in $C$ into $C^{metatrain}$ and $C^{metatest}$.
    \item Generate two (meta-)sets of problems for both meta-training and meta-test stages, $P^{\text{metatrain}}$ and $P^{\text{metatest}}$.
    \item Generate each problem $p_{i}$ in the following way: 
    \begin{enumerate}
        \item Select $K$ classes from either $C^{\text{metatrain}}$ or $C^{\text{metatest}}$.
        \item Generate train and test sets $S_{\text{train}}$ and $S_{\text{test}}$ by picking $N$ and $M$ samples and their respective labels from the $K$ classes. Furthermore, create their unlabeled versions $U_{train}$ and $U_{test}$ (by ignoring their labels). Preserve the test labels $Y_{test}$.
        \item{Specify the budget $B$ and number of classes $K$.}
    \end{enumerate}
\end{enumerate}

This setting allows us to (meta-)train by, at each epoch, randomly selecting a problem from $P^{\text{metatrain}}$, using the model to solve it and evaluating results. Those results are then used to update the model before the new epoch.

At (meta-)test time the process is repeated using the problems in $P_{\text{metatest}}$ set to evaluate the results. This time, the model is not updated.

Note that we have not specified any model or algorithm for any step on the learning problem, we even ignored which kind of update is applied to the model or which results are used to do it. That is because the scope of this section is just to state the scenario to work on.

\subsection{Sample-focused method}
\label{sbsec:prevmethod}

In~\cite{DBLP:journals/corr/ContardoDA17} the model is split in 3 parts: Representation Model (RM), Selection Model (SM) and Prediction Model (PM).

The Representation Model finds a common $F$-feature space to embed samples for both SM and PM, and it is learned through the Meta-training (meta-)stage alongside other components.

The Selection Model gets $U^{R}_{\text{train}}$ and should return $D$ from the Oracle. Actually, \cite{DBLP:journals/corr/ContardoDA17} suggests a slight variation of that, where the SM returns a probability distribution $\alpha$ for selecting the samples in $U^{R}_{\text{train}}$ as a Multinomial distribution where $\sum_{i=1}^{N} \alpha_{i} = 1$. Furthermore, in the  Meta-training (meta-)stage the SM also returns the possible samplings $D_{\alpha}$ given a budget $B$. On the other hand, on the Meta-test (meta-)stage it just returns the $B$ samples with highest probability in $\alpha$, getting $D_{\alpha_{\text{MAX}}}$. In~\cite{DBLP:journals/corr/ContardoDA17} the SM is implemented using a bidirectional GRU ~\cite{ChoMGBSB14} since the input is considered a sequence of samples. We will discuss about the convenience of this choice in the following sections. This SM is learned alongside the RM, through the Learning problems in the Meta-training (meta-)stage. The group of all possible subsamplings can be represented as $D_{\alpha} = \{ D_{\alpha}^{1}, D_{\alpha}^{2}, ... \}$ where each $D_{\alpha}^{j}$ is a single subsampling.

The Prediction Model gets both the represented unlabeled test set $U^{R}_{\text{test}}$ and a labeled set $D_{\alpha}^{j}$ and predicts the classes of the test set $\hat Y_{test}^{j}$. More concretely, the classification of a test sample is performed by assigning the class with the maximal sum of metric similarities between the test sample and the train samples belonging to this class.

Since this PM does not require learning, only the RM and SM are learned, and they are trained jointly during the Meta-training (meta-)stage. The loss which updates the model is a Policy Gradient~\cite{NIPS1999_1713} that explores all the sampling space $D_{\alpha}$ from the distribution $\alpha$ and their rewards (based on the error of their prediction). The use of Policy Gradient to evaluate the whole probability distribution is justified because single samplings are not differentiable and therefore we cannot backpropagate from them. The loss is computed as follows: 

\begin{equation}
\label{eqn:loss}
    \mathcal{L} = \sum_{j}-{\text{log\_prob}(D_{\alpha}^{j}) r^{j}}
\end{equation}

 where $r$ is the reward of a given sampling computed as $r = - d(\hat Y_{\text{test}}, Y_{\text{test}})$ and $d(\hat Y_{\text{test}}, Y_{\text{test}})$ is the error between the prediction $\hat Y_{\text{test}}$ over the true labels $Y_{\text{test}}$. On the other hand,  $\text{log\_prob}(D_{\alpha}^{j}) = \log (\text{prob}(D_{\alpha}^{j}))$ and the probability distribution vector $\alpha$ can determine the probability of a given sampling, i.e. $\text{prob}(D_{\alpha}^{j})$. This $\text{prob}(D_{\alpha}^{j})$ can be computed from the probability of the individual samples  defined as $\alpha_{i}$. For example, for a budget $B=2$, the probability of a sampling of the first and second sample may be ($\alpha_{1} \alpha_{2}$).

This loss only gets high values on high rewards with low probabilities, so it tends to make the most profit on high rewards by penalizing the behavior that gives them a low probability.

\subsection{Limitation of Sample-focused method}
\label{sbsec:limprev}

The previous method is useful to avoid getting undesired samples, such as outliers, that will result in a high error.

\begin{figure}[t]
  \centering
  \medskip
  \subfloat[Selecting both frontier samples the prediction works optimally]{
  \includegraphics[width=0.45\columnwidth]{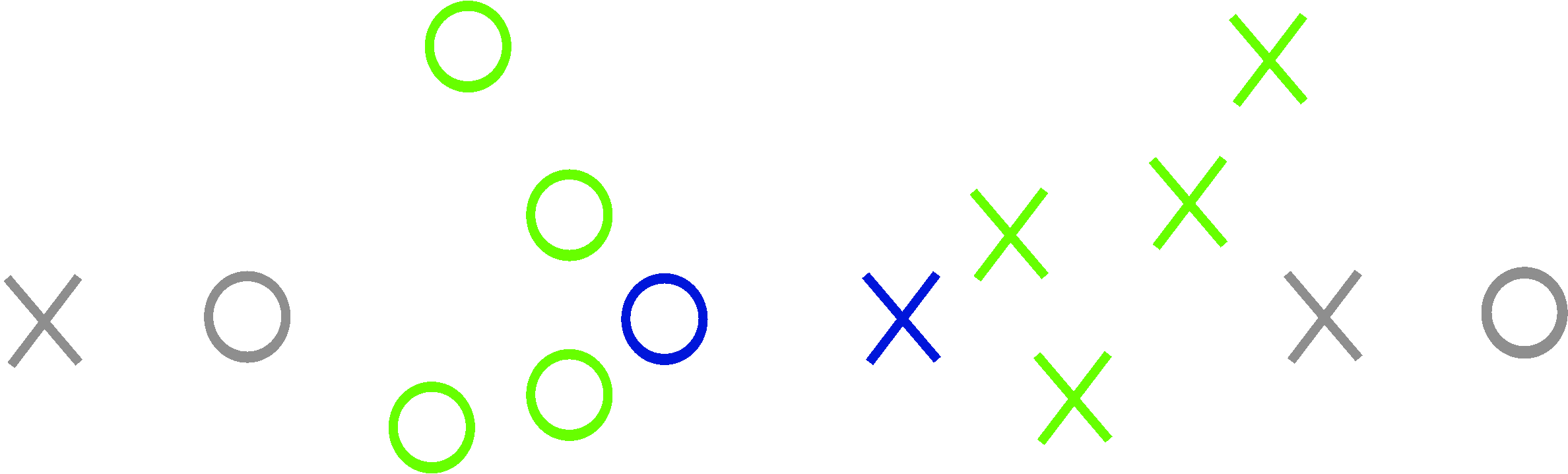}
}
\hfill
\subfloat[When selecting other symmetrical samples the performance is also good]{
  \includegraphics[width=0.45\columnwidth]{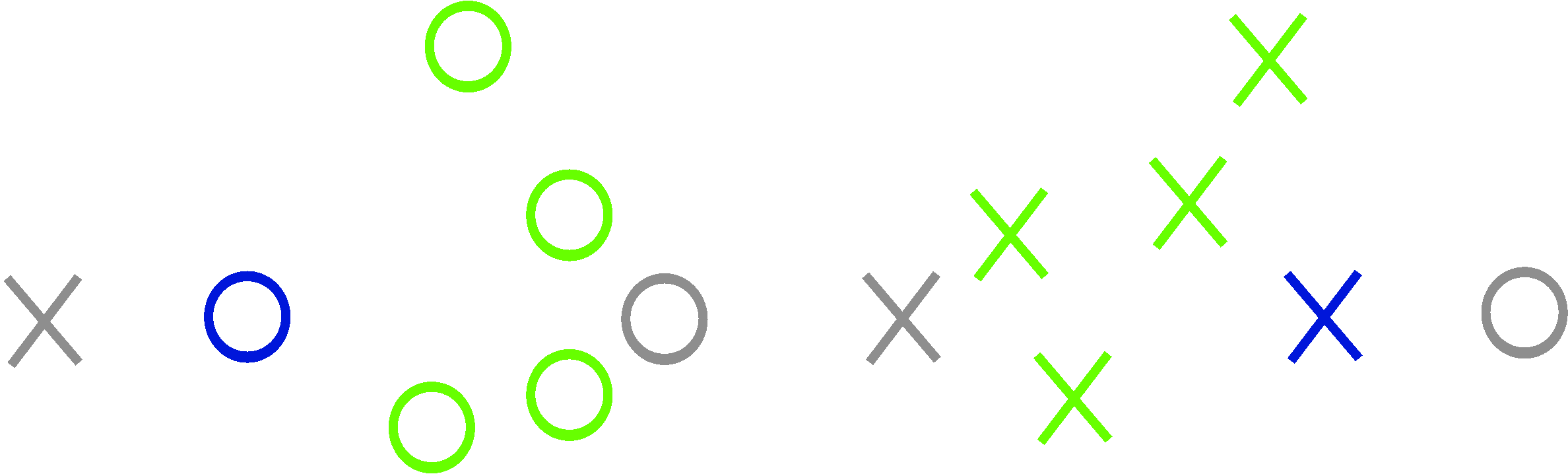}
}
\hfill
\subfloat[Selecting non-equivalent samples for each class is not optimal, even if they work fine in other combinations]{
  \includegraphics[width=0.45\columnwidth]{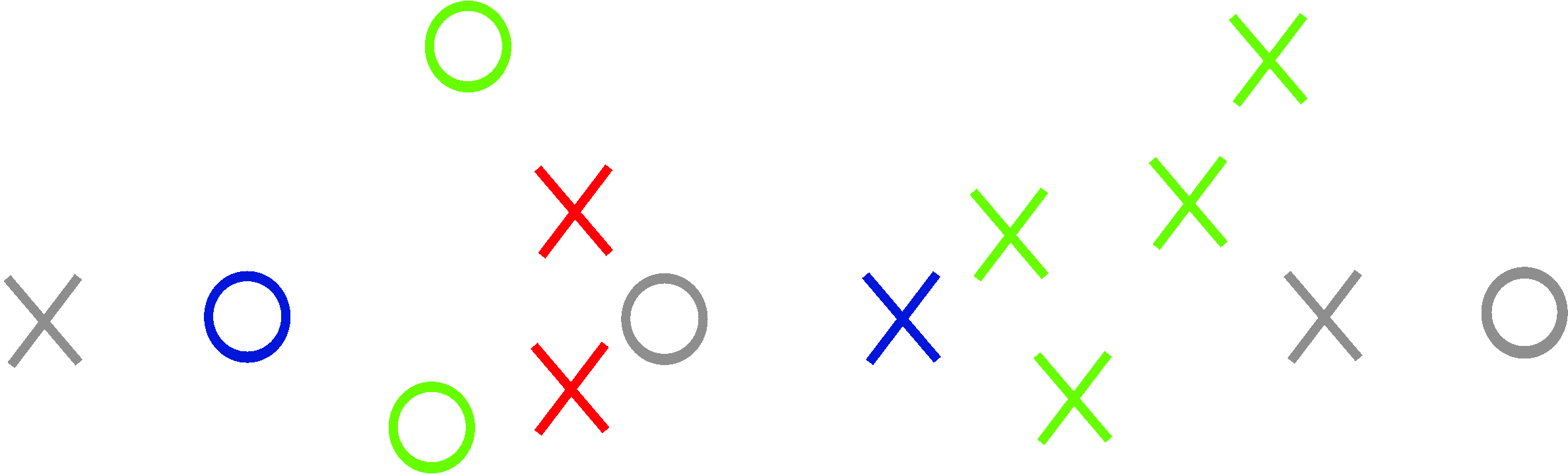}
}
\hfill
\subfloat[Selecting two instances of the same class assigns that class to all the predicted samples which is a really bad performance]{
  \includegraphics[width=0.45\columnwidth]{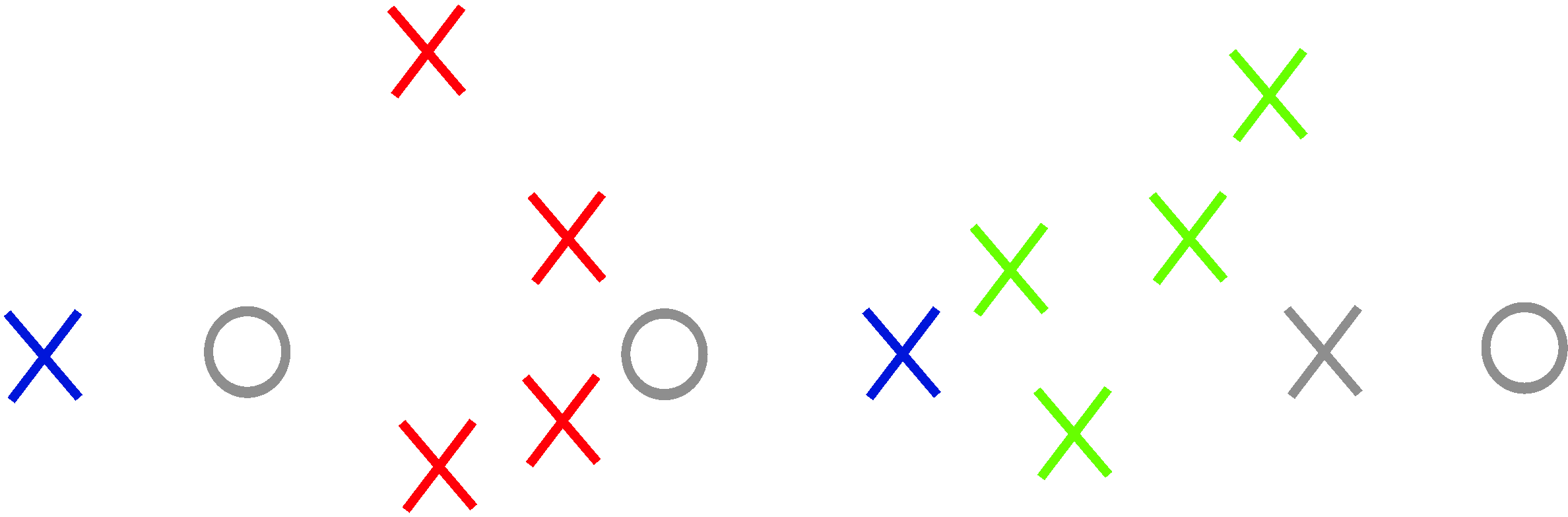}
}
\caption{Example of the effect of each kind of selection using a binary classification problem ($K=2$, illustrated as crosses and circles) with a budget $B=2$. In grey and blue there are the training samples, being blue the selected ones, while green and red are predicted test samples, green for hits and red for misses.}
\label{fig:sel_examples}

\end{figure}

However, it only uses the probability of selecting each individual sample \textit{independently of the rest of the selected samples}. Nevertheless, a single sample can work very differently depending on the rest of the selected samples, as illustrated in \figref{fig:sel_examples}. There is one clear case of these limitations, which is when all the selected samples are from the same class. Even if the samples are very representative of their class, the result will obviously be very bad since all the test samples will be predicted as the same class.

\subsection{Proposed methods}
\label{sbsec:newmethod}

\subsubsection{General architecture}
\label{sbsbsec:methodgeneral}

We defined a method following the same pipeline as in \ref{sbsec:prevmethod}. However, we extended it to overcome the issues mentioned in \ref{sbsec:limprev}. Unlike the Sample-focused method, which just defines a single probability distribution for samples, we propose two ways of handling combinations of samples: the Iterative method and the Combinatorial method.

\begin{itemize}
\item The \textbf{Iterative method} consists of picking the samples one by one until the budget $B$ is met. Moreover, we aggregate to the SM input the information about which samples have already been selected. The idea is to make the selector sensible to the already selected samples at each moment.

\item The \textbf{Combinatorial method} consists of computing a single probability distribution (as in the method in \ref{sbsec:prevmethod}), but unlike the Sample-focused method, this distribution is computed over combinations of samples. The length of the computed probability distribution vector $\alpha$ increases exponentially with the budget $B$.
\end{itemize}

\subsubsection{Representation model}
\label{sbsbsec:methodrm}

We used the convolutional part of VGG-16~\cite{SimonyanZ14a}, pretrained on ImageNet, and we stacked a Fully Connected layer to the desired $F$-feature space. We used $F = 128$.

\subsubsection{Selection model}
\label{sbsbsec:methodsm}

Our main improvements are focused on this component. The goal is to obtain, from an input vector of $N$ samples (vectors of size $F$), the probability distribution vector $\alpha$. 

For the Iterative method, we append to each input sample feature vector an additional binary feature indicating whether it has been already selected or not, thus getting a $(F + 1)$-feature vector for each sample. We propose to scale each embedded feature on the range $[-1, 1]$ and the new binary feature as $\{0, F\}$, getting a vector of $N$ ($F+1$-dimensional) samples.

The Combinatorial method receives $U^{R}_{train}$ and then represents all possible combinations by just stacking the corresponding $B$ samples for that combination, getting a vector of $N^{B}$ ($(B F)$-dimensional) samples. From this vector it needs to return $\alpha$ as a vector of size $N^{B}$, being the probabilities of each combination.

As for the component, we need a model that takes a number of input samples ($N$ vectors of size $F+1$ for the Iterative method and $N^{B}$ vectors of size $B F$) and returns a vector $\alpha$ with the same size as the input, where each element refers to each input element (sample or combination) but taking into account the rest of elements in a sequence to sequence manner~\cite{DBLP:journals/corr/SutskeverVL14}. The model selected for this task is a \textit{Bidirectional GRU} with a stacked Fully Connected layer at the end for the resulting features (as in~\cite{DBLP:journals/corr/ContardoDA17}). We also use Softmax as the activation function. Additionally, the probability of selecting the samples already selected is set to 0.

There is something more we should take into account when generating $\alpha$. The input $U^{R}_{test}$ is actually a set, not a sequence, which means that its samples are ordered randomly and for different problems there is no relationship between their orders. However, Recurrent Neural Networks are designed to handle sequences. To overcome that issue we propose one of the solutions that Vinyals \etal suggest in ~\cite{44871}, which is to force an artificial order. We propose to order the samples according to their distance to a reference. Among different distances and references that have been tested, the one that has given us the best results for the Iterative method is the Euclidean distance to the centroid for the already selected samples. For the Combinatorial method, ordering to the centroid of the already selected samples is not possible (since selection is done in a single step) so we considered ordering by Euclidean distance to the $U^{R}_{test}$ centroid.

\subsubsection{Prediction model}
\label{sbsbsec:methodpm}

We have based the PM on the distances between training and test vectors (in our case, Euclidean distance) as in~\cite{DBLP:journals/corr/ContardoDA17}. More concretely, we have computed the probability to belong to each class as follows:

\begin{equation}
\label{eqn:pm}
    \hat y^{c_{j}}_{test_{i}} = \frac{\sum_{x_{\text{train}_{k}} \in D_{\alpha}, y_{\text{train}_{k}} = c_{j}}{\text{dist}(x_{\text{train}_{k}}, x_{\text{test}_{i}})^{-1}}}{\sum_{l}{\sum_{x_{\text{train}_{k}} \in D_{\alpha}, y_{\text{train}_{k}} = c_{l}}{\text{dist}(x_{\text{train}_{k}}, x_{\text{test}_{i}})^{-1}}}}
\end{equation}

\subsubsection{Loss}
\label{sbsbsec:methodloss}
Policy Gradient requires the exploration of all possible samplings at (meta-)training time. In the Iterative method, the probability distribution is computed for each selected sample in an iterative way (given the previously selected samples). For this reason, if all samplings are explored at each selected sample, a probability distribution needs to be computed for each possible previous sampling. The probability of a final sampling is the product of the probabilities of the selected sample in each selection step in $B$.

We propose to compute the reward as a value that decreases as the error increases in an exponential way, where $r = e^{- d(\hat Y_{\text{test}}, Y_{\text{test}})}$. We use Cross Entropy as the error metric since it is a typical classification loss (so $d(\hat Y_{\text{test}}, Y_{\text{test}})$ is the Cross-Entropy error between $\hat Y_{\text{test}}$ and $Y_{\text{test}}$).

\subsubsection{Training procedure}
\label{sbsbsec:methodtraining}

The procedure to guide the Meta-training consists of computing the loss for each epoch and updating the model with it. Note that the loss does not depend just on the quality of the model but also on the difficulty of the specific random problem, which can undesirably guide the update. To overcome this issue we propose to use several problems per epoch to smooth the randomness of the epoch. We use $50$ problems per epoch. We just need to sum or average the losses of all the problems.

As already mentioned, the model is composed of RM, SM, and PM. However, PM is not trainable, so the optimizer will actually update RM and SM. We propose to pre-train RM for classification problems (through plain Meta-Learning) and then freeze it and just train the SM in the Meta-Active Learning setting.

\section{Experimental validation}
\label{sec:experiments}

\subsection{Experiments setting}
\label{sbsec:expsetting}

Omniglot~\cite{dataset:omniglot} is a dataset of handwritten characters specifically created for Few-Shot Learning. It consists of 50 alphabets.

Each alphabet is a group of characters, which are the actual classes. We split the alphabets into three disjoint (meta-)sets: (meta-)training, (meta-)validation and (meta-)testing alphabets. The method described in~\ref{sbsec:scenario} is used: we construct several Learning problems to guide the (meta-)training of the algorithm and allow a (meta-)evaluation. For each (meta-)set, problems are generated by picking one random alphabet and selecting $K$ random classes from it. The problems consist in classifying samples from these $K$ classes. For each problem $p^s_{i}$, we create the training and test sets with samples from the selected classes.

The key is that any alphabet will not be found in more than one (meta-)set, but the problems will be of similar nature: classification problems of $K$ characters on the same alphabet, but with a random alphabet on each problem.

Using 30 (meta-)training, 10 (meta-)validation and 10 (meta-)test alphabets, we defined a total of 4000 (meta-)training, 500 (meta-)validation and 500 (meta-)test problems of $K=2$ and $B=2$. That means that we are actually building a Meta-Active Learning setting where we want to learn to convert Active Learning problems to Binary One-Shot Learning problems and solve them through Meta-Learning.

Furthermore, we define $N=15$ (15 unlabeled training samples from which we want to label 2 of them) and $M=30$ (30 test samples to predict).

With that setting we compute two metrics: Multi-class ratio, which tells how many problems (relatively) have been solved selecting samples from different classes (a metric that makes sense for binary One-Shot Learning classification problems, i.e. $K=2$, $B=2$) and final accuracy.

\subsection{Results}
\label{sbsec:results}

For fair comparison, we have trained all the approaches on the same benchmark (meta-)training set and evaluated on also the same (meta-)testing set, where the set of (meta-)training and (meta-)testing (and also meta-validation) problems fulfill that $P_{\text{train}} \cap P_{\text{test}} = 0$. The results obtained for the different approaches on the evaluation over the (meta-)testing set are presented in table~\ref{tab:results}.

\begin{table*}[tb]
\centering
\begin{center}
   \begin{tabular}{| c | c | c | c |}
   \hline
           & & Multi-class ratio  & Mean accuracy\\ \cline{1-4}
    \textbf{1} & \textbf{Random selection}  & 0.246 & 0.5385 \\ \cline{1-4}
    \textbf{2} & \textbf{Sample-focused method (previous)} & 0.416 & 0.4981\\ \cline{1-4}
    \textbf{3} & \textbf{Iterative method (unordered)} & 0.278 & 0.5373 \\ \cline{1-4}
    \textbf{4} & \textbf{Iterative method (ordered)} & 0.784 & 0.6473 \\ \cline{1-4}
    \textbf{5} & \textbf{Combinatorial method (ordered)} & 0.374 & 0.5615 \\ \cline{1-4}
    \textbf{6} & \textbf{Best selection} & 0.998 & 0.8675 \\ \cline{1-4}
   \end{tabular}
\end{center}
\caption{Results for each approach, explained in \ref{sbsec:results}}	
\label{tab:results}
\end{table*}

First, as a worst-case scenario, we have tested a system where the Active Learning component selects the samples randomly (row 1). We have also made one experiment evaluating all possible selections (over the samples represented with the pretrained RM) and keeping the best one. The rest of the experiments (corresponding to the proposed approaches) are on this list:

\begin{itemize}
    \item \textbf{Sample-focused method:} We have replicated the method defined by~\cite{DBLP:journals/corr/ContardoDA17}.
    
    The results are presented in row 2. Comparing with the Random Selection (row 1), we see that there is no increase in final accuracy even having a higher Multi-class ratio, which shows that even if we pick samples from different classes we can have bad results (\eg example \textit{c} in \figref{fig:sel_examples}).
    
    \item \textbf{Iterative method, unordered samples:} In this experiment, we evaluate the Iterative method described in~\ref{sbsec:newmethod}, without sample order.
    
    The results are presented in row 3. There is a slight improvement in accuracy. Something also noticeable is the decrease of the Multi-class ratio. This could be caused simply because, without order, the algorithm still does not find a proper behavior.
    
    \item \textbf{Iterative method, ordered samples:} At this point, we study the effects of imposing an order to the input of the SM as described in \ref{sbsec:newmethod}. As already told, this order consists of Euclidean distance to the previously selected samples centroid.
    
    The results are presented in row 4. Here we note a substantial improvement on both Multi-class ratio and on final accuracy. That is because the order helps the SM to better handle the selection depending on how far the samples are from the already selected ones. This ordering gets samples that tend to be from different classes (although not always, it is a difficult unsupervised task).
    
    \item \textbf{Combinatorial method:}  Finally, we have experimented with the method  explained in \ref{sbsec:newmethod}. In this case, we force the order at the input as the Euclidean distance to the pool centroid (there is no other possible origin for this method). 
    
    The results presented in row 5 show a decrease for both the Multi-class and for the final accuracy with respect to the Iterative method with order. However, the reason may be that the order considers the pool centroid as the origin. This ordering has also been used with the Iterative method, with poor results.
    
\end{itemize}

\section{Conclusions}
\label{sec:conclusions}

In this work, we face the problem of Static Pool-based Active Learning as a Meta-Active Learning problem. We present two main contributions, and we improve the previously defined approaches within the same scenario. 

The first one is making the selection Group-focused (within the Static scenario), thus giving the system the capacity to propose optimal groups of samples (i.e. making the selection optimal as a whole). We propose two methods here: the Iterative method and the Combinatorial method. We prove that both outperform the Sample-focused method. 

The second main contribution consists in enforcing an artificial order to the pool of samples. We suggest ordering by distance, to the centroid of either the whole unsupervised pool or the already selected subset as in~\cite{44871}. We prove that the order helps the SM to find some patterns.

In the end, we get some results that encourage us to follow the proposed ideas since we show that they have an effect on the performance in the considered scenario. However, there is still a need to overcome the stated limitations.

The next logical step should be using an Attention RNN (a concept introduced in~\cite{DBLP:journals/corr/VaswaniSPUJGKP17}), which is a more robust approach that can be better than the enforced artificial order.

\section{Acknowledgements}

This work has been developed in the framework of project TEC2016-75976-R, financed by the Spanish Ministerio de Econom\'{i}a y Competitividad and the European Regional Development Fund (ERDF)


{\small
\bibliographystyle{ieee}
\bibliography{bibs.bib}
}

\end{document}